\documentclass{article}
\usepackage[a4paper, total={6in, 8in}]{geometry}

\usepackage[dvipsnames]{xcolor}
\usepackage{graphicx}
\usepackage{enumitem}

\usepackage{float}
\newfloat{Box}{tbp}{lob}
\floatname{Box}{Box}

\begin{document}

\title{Justifying bio-inspired robotics research: \\ 
A taxonomy of strategies}

\author{
  Margaret J. Zhang\ \\ Mechanical Engineering, \\ University of Michigan, USA 48109 \\
  \and
  Justin Ting \\ Electrical and Computer Engineering,\\  University of Michigan, USA 48109
  \and
  Talia Y. Moore \\ Robotics, Mechanical Engineering, \\ 
  Ecology and Evolutionary Biology, Museum of Zoology,\\
  University of Michigan, USA 48109
}

\date{May 2026}

\maketitle

\begin{abstract}
For most of human history, we have not thought systematically about how and why we incorporate aspects of the natural world into our designs. 
The lack of a systematic approach has resulted in inconsistencies in motivations and methods that make it difficult to predict or evaluate the success of bio-inspired design. 
This mismatch between expectations and results can lead to disappointment when a reader considers a bio-inspired design to be superficial, weak, or incomplete. 
This is especially true in the field of Robotics, in which similarity to a biological system might be the driving motivation for construction.
In an effort to assist robotics researchers justify their specific bio-inspired approach and to assist funding program managers with discerning the value of different bio-inspired approaches, here we propose a taxonomy of motivations for bio-inspired design and describe the potential significant contributions that are likely to result from different approaches.
\end{abstract}

\section{Introduction}
The natural world has provided inspiration and mechanistic insight to human engineering for thousands of years \cite{dicks_philosophy_2016}. 
More recently, biological investigation has uncovered novel tasks, mechanistic insights, and even the raw materials to construct designs, establishing itself as an engineering approach. 
Bio-inspiration is also intrinsically tied to the origin of the concept of robotics: the first use of the word ``robot'' in the play \emph{Rossum's Universal Robots} introduced the concept of a synthetic form created in a factory to resemble humans and perform forced labor \cite{capek1920rur}.
Thus, the original concept of a robot was biomimetic.

\begin{figure}[t]
    \centering
    \includegraphics[width=.9\linewidth]{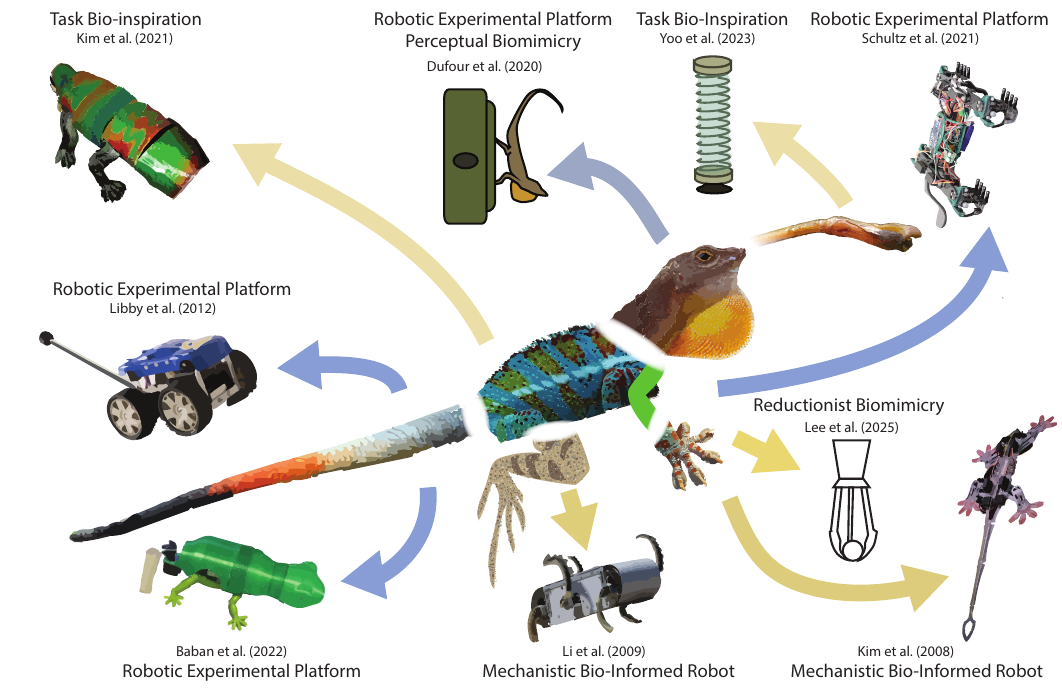}
    \caption{Conceptual diagram showing examples of many different types of bio-inspired design, all inspired by some form of lizard.}
    \label{fig:lizard}
\end{figure}

A wide, multi-faceted, and conflicting array of robot design strategies and approaches wear the ``bio-inspired'' badge.
While ``bio-inspired,'' ``biomimicry,'' ``bionic,'' and ``bio-informed'' are common buzzwords, they are not specific enough describe the diverse ways in which biology is applied to inform design. 
Without clarity, the terms are used imprecisely, especially by the robot engineers who vary in their engagement with biology during their design process. 
At its strongest and most satisfying, a robot may be the result of years of collaboration between engineers and biologists that reveal foundational performance mechanisms across both biology and robotics \cite{habib_bioinspiration_2007, ramdya2023neuromechanics, iida_biologically_2016}. 
At its weakest, a robotics paper may claim to be bio-inspired without referencing biological knowledge, nor justifying why bio-inspiration is necessary or appropriate with respect to the application.
The result of a ``bio-inspired'' robot may also have worse performance than existing, non-bio-inspired examples. 
The dissonance between specific contributions of bio-inspired robotics and the weight of expectations associated with the term has reduced support for bio-inspiration as a design approach.

Although there are ISO standard definitions for biomimicry, they do not provide sufficient granularity to discern among the diverse goals of bio-inspired research \cite{wanieck_perception_2021}.
One review clustered research in biomimicry to consider the common gaps and challenges \cite{sharma_biomimicry_2019}. 
There are also papers that provide guidance on rigorous and principled approaches to perform bio-inspired design, which have only recently been established \cite{koditschek_principled_2004, fu_bio-inspired_2014}. 
A previous review provided a system for determining the depth of the biological analogy, which is useful especially when considering how to strengthen your bio-inspired design \cite{harvey2026bio}. 
A reliable way to strengthen the analogy in your design process is to work directly with biologists, as described in another paper \cite{barley_addressing_2022}. 
However, they also point out that siloed training and a mismatch of research goals are often barriers to establishing such interdisciplinary relationships. 
While many of these reviews have conflicting definitions of related terms, there is a general consensus that categorization is useful for distinguishing different bio-inspired approaches.

The goal of this paper is to categorize the range of contexts in which the term ``bio-inspired robotics'' has been applied in a way that enables researchers to justify their approaches, clarify their goals, and set reasonable expectations for the outcomes of their work.
We introduce specific categories that better reflect the increased diversity of work currently published under this label and map them based on their contributions to science versus engineering and their engagement with biology. 
While the categories presented here are each broad in their own right, they are clearly distinguishable from one another in their approaches and goals, which we hope will result in meaningful differences in a reader’s expectations for contributions in each category. 
More precise language will help researchers identify the context for their own work, ensure it reaches the most relevant communities, and allow those communities to more easily find and build upon it. 
All fields that intersect with bio-inspired research stand to benefit from this clarification. 
Our aim is not to provide a brief history of each category, but to crystallize the differences in goals and approaches among categories to aid researchers in justifying their bio-inspired approach and setting clear expectations for their work in grant proposals, papers, and presentations.

\section{Bio-inspiration Taxonomy}

In particular, we will be focusing on categorizing bio-inspired robotic contributions, but we will include some closely related work, such as algorithms and kinetic art.
We focus on robotics because of the many opportunities for analogy across scale, integration, embodiment, and interaction with the environment. 
Robots have served as a means to test biological hypotheses and explore the function of biological phenotypes, but they also provide an embodied hierarchical system for transferring many different aspects of biological knowledge into engineering. 

We will also provide advice for each category on where to seek funding, where to present the work, and where to publish the findings. 
Due to the location of the authors, the funding advice is primarily based in the USA, but we expect that the advice is transferable because equivalent funding agencies in each country will likely be similarly organized (e.g. science versus engineering research).

\begin{figure}[t]
    \centering
    \includegraphics[width=\linewidth]{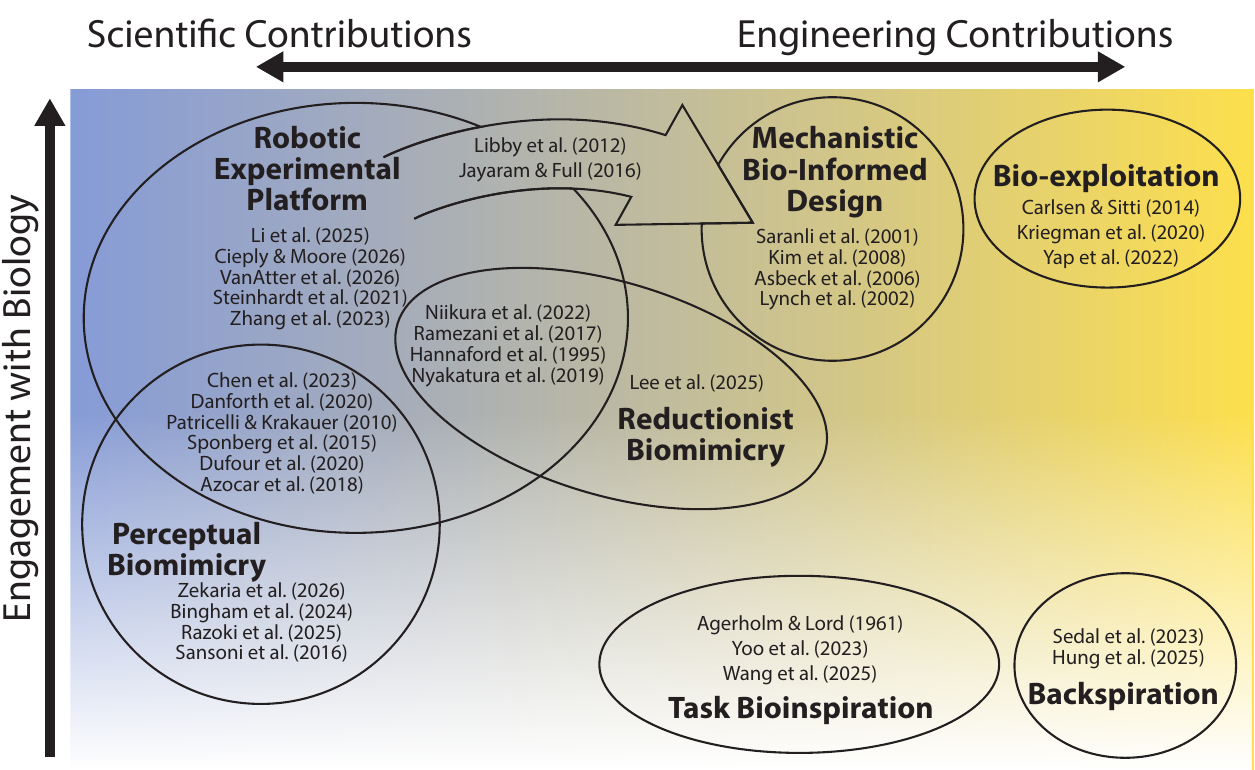}
    \caption{Diagram showing the spatial relationships among the taxonomical categories arranged by Engagement with Biology on the y-axis and contributions to science or engineering on the x-axis. 
    \textit{Disclaimer: This taxonomy is meant to generalize common themes seen throughout the robotics community. 
    There will be unique cases that do not easily classify into these groups.}}
    \label{fig:diagram}
\end{figure}
\subsection{Task Bio-inspiration: Seeking to achieve similar tasks as biology}
\label{sec:task}

Task bio-inspiration describes work in which a biological behavior or capability serves as the motivating goal of a design, without drawing from the underlying biological mechanism used to achieve it. 
The engineer observes that biology successfully accomplishes a given task and adopts that task as a target, but develops an independent mechanism to achieve it \cite{wang2025spirobs,kim2021biomimetic,agerholm1961artificial}. 
In this category, biology mainly serves as an existence proof that a given task is possible.
This distinguishes task bio-inspiration from more mechanistically grounded approaches: the biological system informs the goal, not the how. 
Work in this category does not always explicitly use the term ``bio-inspired,'' yet it represents a meaningful mode of learning from biology and is regularly published in robotics journals (e.g., 
\emph{Soft Robotics}, 
\emph{IEEE Transactions on Robotics}, 
\emph{Robotics and Automation Letters} \cite{yoo2023compliant}, 
\emph{International Journal of Robotics Research}) and presented at robotics conferences (e.g., IEEE RoboSoft, IEEE International Conference of Robotics and Automation, IEEE/RSJ International Conference on Intelligent Robots and Systems, IEEE/ASME Transactions on Mechatronics, IEEE Robotics \& Automation Magazine, Robotics: Science and Systems. 
A key contribution of this approach is that it expands the space of engineering goals by drawing attention to tasks that biology has already demonstrated to be achievable \cite{spot, hutter2016anymal, haldane2016robotic}.

\subsection{Mechanistic Bio-inspiration: Robots that are designed from biological mechanisms}
\label{sec:mech}

Mechanistic bio-informed design translates well-characterized biological mechanisms and physical principles into engineered systems to enhance performance or confer new capabilities. 
A distinguishing feature of this approach is that the biological mechanism must be understood and abstracted so that it can be transferred to design. 
This makes mechanistic bio-informed design a more rigorous and biologically grounded form of bioinspired design, as the biological basis for each design decision is traceable to specific curiosity-driven experimental research. 
Biological mechanisms have inspired the design of sensors \cite{Event_Based_Survey}, actuators \cite{agerholm1961artificial}, robot morphology \cite{jayaram2016cockroaches, saranli2001rhex}, material properties \cite{li2009sensitive}, and controllers for locomotion \cite{ramdya2023neuromechanics, koditschek_principled_2004} and other behaviors \cite{wong2011bioinspired}.
In most cases, the engineer draws upon the description of biological mechanisms from previously published work. 
However, there are also cases in which discovery and design are performed by the same team of interdisciplinary researchers and published in the same manuscript \cite{Libby2012, ijspeert2007swimming,jayaram2016cockroaches}. 

There is a distinct advantage to a Mechanistic Bio-Informed approach. 
Engineered designs are not subject to evolutionary, genetic, multi-functionality, or developmental constraints that may prevent biological systems from achieving global performance optima. 
Therefore, Mechanistic Bio-Informed systems can actually exceed the performance of the biological analog \cite{white_tunabot_2021,Libby2012}.

This work is published primarily in robotics journals (listed in \textbf{Task Bio-Inspiration}) \cite{asbeck2006climbing,kim2008smooth,lynch2022autonomous},
as well as more interdisciplinary journals, such as \emph{Bioinspiration \& Biomimetics} and the \emph{Journal of the Royal Society: Interface}. 
In addition to robotics conferences (listed in \textbf{Task Bio-Inspiration}) \cite{kim2008smooth}, the work can be presented at the Adaptive Motion of Animals and Machines and the Living Machines conferences.
Because the ultimate goal is to build a better design by leveraging biological insights, this work can be funded by Engineering programs like the National Science Foundation's programs on Foundational Research in Robotics (FRR), Engineering Design and Systems Engineering (EDSE), or Dynamics, Control and Systems Diagnostics (DCSD). 
This work also fits well with calls for translational research.

\subsection{Reductionist Biomimicry: Copying the details of biology without big picture function}
\label{sec:copycat}

Reductionist Biomimicry describes work in which the morphology or details of a biological system are reproduced as faithfully as possible.
This is similar to the concept of a Digital Twin, but in physical form.
The emphasis is on biofidelity rather than functional equivalence \cite{lee2025grip}; the designer replicates the form of the biological counterpart, but not necessarily the mechanisms that allow it to perform \cite{ramezani2017biomimetic}. 
In many cases, the high degree of biofidelity can provide useful models for understanding how one specific organism or system functions, an approach that has been called ``Deep Biomimicry '' \cite{nyakatura2019reverse, hannaford1995anthroform,niikura2022giraffe}.
Similarly, Reductionist Biomimicry performed iteratively can help reveal which structures are essential for the operation of a system, therefore overlapping with the \textbf{Robotic Experimental Platform} category.
Hypothesis-driven Reductionist Biomimicry therefore aids in providing specific insights on a narrow range of biological subjects, usually into the combined coordination and operation of components.
This type of work is often published in interdisciplinary journals, such as \emph{Bioinspiration \& Biomimetics} and \emph{Journal of the Royal Society: Interface}, and presented at scientific conferences, such as the International Congress of Vertebrate Morphology, Society for Integrative and Comparative Biology, American Society of Biomechanics, International Congress for Neuroethology.
There are also interdiscplinary conferences, such as Living Machines and the International Symposium on Adaptive Motion in Animals and Machines at which this work may be relevant.
Depending on the scale and degree of biofidelity, hypothesis-driven Reductionist Biomimicry research could potentially be funded by the National Science Foundation Integrative Organismal Systems program on Physiological and Biomechanical Mechanisms or through the National Institutes of Health's National Institute of Biomedical Imaging and Bioengineering (NIBIB).

A key limitation of this approach is that it does not generalize well: 
biological systems are incredibly complex, and without explicitly identifying the mechanism of function, the design cannot be readily scaled, transferred to new applications, or adapted to different functional contexts. 
At its weakest, a Reductionist Biomimetic design is the result of arbitrary decisions on which features to copy.
Because Reductionist Biomimicry is constrained to the biology, and biology is subject to the myriad constraints described in the \textbf{Mechanistic Bio-Informed Design}, such designs typically exhibit reduced performance compared to their biological analogues.
As such, Reductionist Biomimicry that is not hypothesis-driven primarily contributes to engineering and can be published and presented in the venues listed in \textbf{Task Bio-Inspiration}.
Reductionist Biomimicry can serve as a useful starting point for understanding a biological system, but its contributions to the broader field are limited without a corresponding effort to understand function.

%  REDUCTIONIST
\subsection{Perceptual Biomimicry: Seeking to resemble biology }
Perceptual Biomimicry refers to work in which the biological presentation of a design is the primary objective, rather than the replication of biological function or mechanism. 
The robot can exhibit visual, auditory, odor, or tactile cues perceived by an observer.
A design is successful when it is perceived as biological or naturalistic by its intended audience, making the presentation the function of the design.
This makes Perceptual Biomimicry one of the broadest categories discussed here, as it overlaps significantly with fields beyond robotics, including behavioral ecology \cite{danforth2020emulating, chen_feed_2023, dufour_recent_2020}, architecture \cite{razoki2025biomimetic}, art and entertainment (e.g., animatronics) \cite{bingham2024biomimetic}, and medical technology (e.g., robotic biological phantoms for training medical students or prostheses that must match natural gait patterns and appearances) \cite{pelvic,azocar2018design, sansoni2016aesthetic}. 
Work in this area is published in a correspondingly wide range of venues, including \emph{Bioinspiration \& Biomimetics}, biology journals, and robotics journals \cite{danforth2020emulating}, depending on the application. 
Work on robot designs are typically presented at robotics conferences (discussed in \textbf{Task Bio-Inspiration}) \cite{berrybot}, while research on audience perception and response is better suited to discipline-specific venues in the relevant field.
For example, ACM/IEEE International Conference on Human-Robot Interaction (HRI) and ACM Special Interest Group on Computer Graphics and Interactive Techniques (SIGGRAPH) would both be appropriate for animatronics.
Perceptual Biomimetic robots interacting with animals can be presented at biological conferences, including (but not limited to) the Society for Integrative and Comparative Biology, Animal Behavior Society, or International Society for Behavioral Ecology Congress. 
Funding for this work is generally sought from non-robotics sources, reflecting the interdisciplinary nature of the category.

\subsection{Robotic Experimental Platform: Developing new design approaches based on biological mechanisms}
\label{sec:exp}
Bio-inspired robots can be developed for scientists to enable controlled and isolated experimentation. 
Rather than using live animals or biological specimens directly, a robotic or mechanical surrogate is employed to isolate and study specific properties of a system, such as morphology, kinematics, or behavioral strategies, under repeatable conditions \cite{gravish_robotics-inspired_2018, lauder_robotics_2022, webb_biorobotics_2001, roberts_testing_2014, Libby2012,berrybot}. 
This type of robot can also be called a Robotic Model Organism  if the application is biology, a robotic phantom in medicine, or Robo-Physical Model more generally \cite{flammang_bioinspired_2022, holmes_dynamics_2006}. 
This approach is used across disciplines including biology (often using the term Biorobotics), physics, and medicine to investigate behavior \cite{patricelli2010tactical, dufour_recent_2020}, evolution \cite{danforth2020emulating}, form-function relationships \cite{steinhardt2021physical, Cieply2026, zhang2023launching}, biophysical mechanisms \cite{Sponberg2015InsectFlight}, and to train doctors \cite{roberts_testing_2014, tamborini_is_2022, iida_biologically_2016, white_tunabot_2021}. 
Three key benefits of this approach include 
a) being able to isolate variation in variables that are coupled in the biological system, and 
b) embodying extinct or theoretical parameter values to explore whether an empty area of morphospace is due to biomechanical constraint or chance \cite{nyakatura2019reverse},
c) reducing discomfort for the replaced live organism/patient. 

Depending on the target application, this work spans multiple communities: Biology-focused research with robotic experimental platforms are often published in the \emph{Journal of Experimental Biology}, \emph{Integrative and Comparative Biology}, \emph{Integrative Organismal Biology}, \emph{Behavioral Ecology}, \emph{Proceedings of the Royal Society: B} \cite{schultz2021climbing}, \emph{Bioinspiration \& Biomimetics} \cite{baban2022biomimicking}, and the robotics-for-science track of \emph{Science Robotics} \cite{li2025reverse}. 
Biology research leveraging Robotic Experimental platforms is commonly presented at non-robotics scientific conferences, such as the Society for Integrative and Comparative Biology, the Society for Experimental Biology, the Animal Behavior Society, or the International Congress of Neuroethology. 
The work can also be presented at the Adaptive Motion of Animals and Machines and the Living Machines conferences.
Funding for robotics-enabled biology is most appropriately sought from biology-focused agencies, including the NSF Integrative Organismal Systems (IOS) division, especially the Program for Physiological and Structural Systems (PSS). 
When the primary contribution is to physics, the work is commonly presented at the American Physical Society (APS) March Meeting with funding sought from the NSF Physics of Living Systems (POLS) program. 
Physics insights leveraging Robotic Experimental Platforms are often  published in APS family of journals: Physical Review Letters (PRL), Physical Review E, and Physical Review Fluids.
When the primary application is robotic phantoms to train doctors and surgical robots, the work can be presented and published in several journals, including \emph{Physics in Medicine \& Biology (PMB)}, \emph{Medical Physics}, \emph{Physical and Engineering Sciences in Medicine}, \emph{Journal of Medical Robotics Research}, \emph{IEEE Transactions on Information Technology in Biomedicine}, \emph{IEEE Journal of Translational Engineering in Health and Medicine}.
Robotic phantoms can be presented at several conferences, including the IEEE RoboSoft conference and the IEEE International Conference on Biomedical Robotics and Biomechatronics (BioRob). 
The NIH NIBIB (National Institute of Biomedical Imaging and Bioengineering) has previously funded work on robotic medical phantoms.

\subsection{Bioexploitation: Using biological structures as engineering components}
\label{sec:exploit}
Bioexploitation represents an emerging area of research that is often compared to bio-inspired design because of its direct connection to biology \cite{raman_biofabrication_2024, burden_why_2024, sato_cyborg_2008}. 
However, these fields represent distinct approaches.  
Bio-inspiration draws on mechanistic principles learned from nature to inform design, whereas bio-exploitative work directly incorporates biological material---such as living tissue or deceased animal appendages---into engineered systems \cite{cvetkovic_three-dimensionally_2014, raman_modular_2017, filippi_multicellular_2025,carlsen2014bio}. 
There are two distinct advantages to a bio-exploitative approach: 
a) Biology manufactures at a much smaller scale than is currently possible with human manufacturing methods, so co-opting biological structures allows humans to operate at a smaller scale, and 
b) Biological system components are better integrated than manufactured systems, allowing engineers to use pre-integrated biological structures to replace a whole suite of sensors, springs, motors, and pumps. 
While bioexploitative work does engage with biology in a direct and meaningful way, it has grown sufficiently distinct in its methodology and goals to warrant its own classification rather than being grouped under bio-inspiration \cite{mishra_sensorimotor_2024, dogan_immune_2024}. 

The Bioexploitation field contains more specific sub-categories that reflect the nature of the biological material used. 
For example, ``bio-hybrid'' typically refers to structures composed of grown biological tissue that perform biological functions while cells are alive.
Bio-hybrid tissues have successfully served as actuators, sensors, controllers, or entire robots \cite{mishra_sensorimotor_2024,kriegman2020scalable}.
On the other hand, ``Necrobotics'' refers to work in which components from deceased organisms are integrated into robotic systems \cite{yap_necrobotics_2022, puma_3d_2025, kim_dead_2026}.
Cyborgs are created by implanting electronics into an insect during development, allowing engineers to control the behaviors of the adult \cite{sato_cyborg_2008}.
All three approaches leverage the existing capabilities of biological tissues by physically incorporating the tissues themselves into a robot.
We recommend reserving the term ``bio-inspiration'' for work in which biology informs the engineering process conceptually rather than materially. 

The field of bio-hybrid research has rapidly grown and is generally published in \emph{Advanced Materials}, \emph{Advanced Healthcare Materials}, \emph{Biotechnology and Bioengineering}.
and presented at the Living Machines conference as well as the robotics conferences listed in \textbf{Task Bio-inspiration: Seeking to achieve similar tasks as
biology}.
Several NIH institutes (NIBIB, NICHD, NINDS, NCI) could fund biohybrid research, depending on the application.
Research on cyborgs has been published in \emph{Advanced Science} and \emph{Advanced Intelligent Systems}, and presented at the IEEE International Conference on Micro Electro Mechanical Systems and the IEEE Transactions on Medical Robotics and Bionics.
At the time of writing, Necrobotics is in the early stages of becoming a cohesive field of research. 
Such work has been published primarily in \emph{Advanced Science} \cite{yap_necrobotics_2022,kim2026dead}
and would likely be presented at conferences similar to IEEE RoboSoft and Living Machines.

\subsection{Backspiration: Biological similarity, but not inspirations}
\label{sec:back}
In the loosest application of bio-inspiration, we propose the term ``Backspiration'' to describe work in which the label ``bio-inspired'' is applied retroactively, rather than arising from the design process itself. 
Backspiration can take many forms.
In the most generous interpretation, an author invokes a biological analogue as a post-hoc shorthand for contextualizing a gap in the field that the project addresses, but the analogy between the inspiration and the project is weak.
In other work, the term functions as a keyword appended to increase publication visibility, regardless of whether biology meaningfully informed the design. 
This latter practice can be harmful to the research community in two ways. 
First, it oversaturates the literature, making it harder to identify work in which biological principles substantively shaped engineering decisions, as opposed to loose inspiration. 
Second, it erodes the integrity of bioinspired design by creating a gap between what the term implies and what the work delivers. 

To provide specific examples of Backspiration to address a research gap, we point to two works from our own research group---SKOOTR \cite{hung2025skootr} and the Sequential Auxetic robot \cite{sedal2023emergent}---in which biology was considered only after the design and fabrication of each system were complete. 
SKOOTR was vaguely bio-inspired at first, and we therefore largely omitted the term ``bio-inspired'' from the first manuscript draft.
However, in response to reviewer comments, we identified specific biological organisms that exhibit an equivalent form-function relationship to illustrate the novelty of the robot design.
At the time of writing the Sequential Auxetic paper, we used a biological example as an example of the desired goal: time-delays between motions in a connected volume without valves. 
The absence of biological reasoning at the outset of the design process is precisely what distinguishes Backspiration from the other categories of bioinspiration. 

Backspiration is most commonly found in robotics journals and conference proceedings (listed in \textbf{Task Bio-inspiration: Seeking to achieve similar tasks as
biology}).
However, we discourage Backspiration as an approach to bio-inspired research, as it makes it more difficult for engineers to genuinely engage with biology in the design process. 
Indeed, funding agencies deciding on which areas of research to invest may become disillusioned with the entire concept of Bio-inspired Design if the field is saturated with Backspiration.

\subsection{Dichotomous Key}

To guide researchers in identifying the most appropriate category for a given research topic, we introduce a classification tool drawn from biology itself: a dichotomous key (Box \ref{box:key}). 
A dichotomous key is a sequential decision tool in which the user answers a series of binary questions in order; each answer leads the user to a narrower set of conditions until the final classification is met. 
One important difference between a rigorous dichotomous key and the classification tool presented here is that bio-inspired projects can be part of multiple categories. 
So, instead of simply stopping at the first classification that fits, we encourage users to explore both pathways if a project spans multiple answers to a query.

\begin{Box}[t]
    \centering
    \fbox{
        \begin{minipage}{0.9\linewidth}
        \begin{enumerate}[label=\arabic*.]
            \item Did you build the robot before thinking about biology?
            \begin{enumerate}[label=\alph*.]
                \item Yes. \textbf{Backspiration}
                \item No. Go to 2.
            \end{enumerate}
            \item Do you have a specific scientific hypothesis to test?
            \begin{enumerate}[label=\alph*.]
                \item Yes. \textbf{Robotic Experimental Platform}
                \item No. Go to 3.
            \end{enumerate}
            \item Are you using biological components?
            \begin{enumerate}[label=\alph*.]
                \item Yes. \textbf{Bio-Exploitation}
                \item No. Go to 4.
            \end{enumerate}
            \item Do you understand the physics of a specific biological mechanism?
            \begin{enumerate}[label=\alph*.]
                \item Yes. \textbf{Mechanistic Bio-informed Design}
                \item No. Go to 5.
            \end{enumerate}
            \item Must you consider how your design will be perceived by an observer?
            \begin{enumerate}[label=\alph*.]
                \item Yes. \textbf{Perceptual Biomimicry}
                \item No. Go to 6.
            \end{enumerate}
            \item Are you emulating biological structures?
            \begin{enumerate}[label=\alph*.]
                \item Yes. \textbf{Reductionist Biomimicry}
                \item No. \textbf{Task Biomimicry}
            \end{enumerate}
        \end{enumerate}
        \end{minipage}
    }
    \caption{A relaxed dichotomous key to aid in categorizing a robotics project into a specific form of bio-inspired design.
    An important difference from traditional dichotomous keys is that a bio-inspired robotics project can be in multiple categories.}
    \label{box:key}
\end{Box}

\section{Discussion}
In this section, we demonstrate the taxonomy’s flexibility to classify research and the downstream implications of those classifications. 
Not only do we address potential confusion for some un-intuitive cases, but we also clarify the boundaries of each category. 
We hope the proposed taxonomy enables researchers to invoke the term ``bio-inspired'' with meaning and substance, as well as provide a basis for thought-provoking conversations that advance our collective understanding.

\subsection{Overlapping categories}
Robotics projects frequently occupy more than one category (Fig. \ref{fig:diagram}). 
As described in \textbf{Reductionist Biomimicry: Copying the details of biology
without big picture function}, Reductionist robotic models that mimic biological structures can become Robotic Experimental Platforms when they are used to test hypotheses regarding coordination and integration \cite{hannaford1995anthroform}.
Similarly, robots that are developed for animal interaction experiments in behavioral ecology are both Robotic Experimental Platforms and Perceptual Biomimicry \cite{danforth2020emulating,chen_feed_2023}. 

On the other hand, even for the subset of systems in which structure does confer function, Reductionist Biomimicry does not overlap with Mechanistic Bio-Informed Design due to the lack of identifying the mechanistic principle that governs the function.
Backspiration and Reductionist Biomimicry cannot overlap because the first approach references biology as an afterthought, while the latter closely mirrors components of the biological system.
Task Bioinspiration does not overlap with Reductionist Biomimicry nor with Mechanistic Bio-Informed Design because it only considers the end goal, without considering the biological solution to the problem.
Although being perceived could be considered a task, Task Bio-inspiration and Perceptual Biomimicry do not overlap because the perceived robotically generated signal must closely match the biological signal, whereas a design inspired by a biologically performed task does not share characteristics with the biological system performing the task.

Projects can also shift from one category to another.
For example, Robotic Experimental Platforms frequently spark Mechanistic Bio-Informed Design strategies (arrow in Fig. \ref{fig:diagram}).
In some cases, the same research publication contributes to both categories \cite{libby_tail-assisted_2012,jayaram2016cockroaches}. 
While successful designs may demonstrate and validate a biological mechanism, biological knowledge cannot be extracted from the design process alone. 
In other words, knowledge flows only in one direction between these two categories.

\subsection{Bio-inspired Spinoffs and Compound Analogies}\label{discuss:spinoffs}
Bio-Inspired Spinoffs are second-generation technologies that leverage a previously bio-inspired component. 
Spinoffs applying the existing bio-inspired mechanism in a new context are not grandfathered into the bio-inspired category. 
For example, consider a bio-inspired suction cup that is duplicated and placed on a wheeled robot.
While the suction cup was bio-inspired, the wheeled robot is not.
Similarly, algorithms that are tailored towards spike-based bio-inspired hardware, such as visual odometry for event-based cameras \cite{Event_VIO}, are also spinoffs. 
Despite the event-based visual system being inspired by nature \cite{Event_Based_Survey,Mahowald1994}, an algorithm tailored to this type of sensor does not need to be inspired by how a brain would process visual stimuli. 

On the other hand, if a new source of biological inspiration is used to add a feature to an existing bio-inspired design, it can become a compound analogy.
For example, a robot body inspired by the material properties, morphology, and gait of a cockroach would become a compound analogy when a lizard-inspired tail is attached to induce precise turns \cite{kohut2013precise}.

During the design process, an idea can begin with bio-inspiration and then move away from the biological analogy.
In such cases, it is valid to leave bio-inspiration out of the final publication.
For example, a lamprey-inspired suction cup began as a team project in a bio-inspired design course.
As the students iterated on the design after the course, bio-inspired features were removed and non-bio-inspired features were added to enhance performance and accomplish novel tasks. Thus the original biological inspiration was removed from the publication \cite{bu2025release}.
One need not feel obligated to maintain a biological analogy if it does not enhance performance.

\subsection{Bio-inspired Algorithms}
\label{subsec:discuss_task}

There are classes of algorithms that are considered bio-inspired, such as neural networks, neuro-inspired AI \cite{NeuroInspiredAI}, evolutionary algorithms \cite{evolutionary_algorithms}, and bio-inspired control theory \cite{Todorov_iLQR, Todorov_motor_theory}.
These algorithms have two bio-inspired components: cost functions and operations. 
The cost functions represent biological phenomena (e.g. evolutionary selection) or behaviors and tasks exhibited by animal nervous systems (e.g. image labeling, path planning). 
Cost functions differ fundamentally from biology because biology does not ``optimize,'' so we classify them as Task Bio-inspiration.
The operations are interpretations of biological processes.
For example, matrix multiplication is an interpretation of neuron integrate-and-fire, and bit flips are an interpretation of genetic mutation, but these inspired operations are not designed to be faithful models of biological processes. 
For this reason, we do not classify this as Mechanistic Bio-inspiration, but rather as Task Bio-Inspiration.
In general, bridging computation with biological reality presents difficulties outside the scope of this review \cite{WhenDoesPhysicalSystemCompute, neuroscience_microprocessor, Poggio2012LevelsOfUnderstanding, MetaphorBrainsAsComputers,digital_sphinx,HuaiTi}.

\subsection{Neuromorphic Robotics}
Most of the examples provided above manifest as mechanical structures or control strategies, but biological inspiration can also extend to the structure of the computers themselves.
Because neural spiking signals use less power than digital signals, the primary motivation for neuromorphic hardware (also called brain-inspired or neuro-inspired) is energy-efficient computation. 
The most notable case of neuromorphic robotics is the application of event-based cameras \cite{Event_Based_Survey,Event_Control_Quad,Event_VIO, event_obstacle_avoidance}, an example of Mechanistic Bio-Informed Design. 
The light sensor of event-based cameras, which is a semiconductor substrate, emulates how the human retina asynchronously samples light intensity \cite{carver_mead_book,Mahowald1994}.

Unlike event-based vision, most biological inspiration for neuromorphic hardware is isolated to the spiking dynamics of single neuron
(Mechanistic Bio-Informed Design) \cite{NeuromorphicSiliconNeuron2011, neuro_memristors, neuro_photonics, neuro_magnets}.
However, these neural units must be aggregated, on the order of $10^2$ to $10^9$ neurons, to execute complex control, estimation, and learning on robots \cite{neuromorphic_PID, neuromorphic_MPC, SNN_SLAM, Zanatta2024ExploringSNNDRLRobotics}. 
Aggregating neurons to implement these algorithms requires a large stack of spinoff technologies (defined in \textbf{Bio-inspired Spinoffs and Compound Analogies}), which includes interconnect, architecture, read-write circuits, encoding, and compilers \cite{nengo, Loihi_compiler, Boahen2000AERPointToPoint, Loihi, BrainScale2, TrueNorth, SpiNNaker2}. 
These spinoffs are the products of computer engineering and the legacy of its tools and methods \cite{hooker2020hardwarelottery}, rather than additional insight from neuroscience. 
Although the distributed and parallel architectures in neuromorphic hardware may seem more brain-like compared to CPUs, these approaches have a long history without bio-inspiration \cite{why_systolic,PIM}.

\subsection{Opportunities for bio-inspired robotics strategies}
\label{sec:conclusion}
The ways in which robotics can be inspired by nature are as multifarious as the muses nature provides.
With our taxonomy and diagram, we recognize that successful robotic systems can arise from these many different approaches to bio-inspired research.
We have found that our categorization scheme has helped us provide tailored feedback as we review papers and proposals.
Below, we describe some category-specific strategies to strengthen the impact of a bio-inspired project.

%Task
\textbf{Task Bio-Inspiration:}
This approach relies on the ingenuity of the designer for the success of the project, as the biology is only serving to demonstrate feasibility.
We recommend that engineers using this approach seek out the least similar to human capability and most extreme tasks performed by biology, which can be selected by partnering with a biologist.

%Reductionist
\textbf{Reductionist Biomimicry:}
Reductionist Biomimicry is strongest when it is used as an experimental platform.
Just as Digital Twins are detailed models that grapple with the complexity and interconnectedness of constituent parts, Reductionist Biomimicry robots could provide similar insights.
Such models can provide insight regarding emergent properties of complex systems.

%Perceptual
\textbf{Perceptual Biomimicry:}
Robotic platforms have immense potential for precisely varying multi-modal signals \cite{partan2013ten}.
We encourage researchers to seek out stimuli outside of our human perception capabilities.
For example, some snakes have pigments that reflect UV light \cite{crowell2024ecological}, spiders sense by inducing vibrations in their webs \cite{barth1998vibrational}, and many animals combine odor cues with motion for communication \cite{vickers2000mechanisms}.
On the other hand, if the observer responds to a unimodal stimulus, consider streamlining your model and removing unnecessary features.
As an extreme example, researchers have used a laser or hologram to provide visual stimuli---of both prey and conspecifics---to fish \cite{ioannou2012predatory,li2025reverse}.

%Mechanistic Design: no notes. keep being excellent.
\textbf{Mechanistic Bio-Informed Design:}
This approach is becoming well-established, with its own structured set of guidelines.
The current popularity of learning algorithms calls into question the value in finding inspiration from nature to develop robot control algorithms.
The counterpart to an entirely learned approach is mechanical intelligence, which leverages the structured compliance of morphology to passively perform behaviors, thereby transferring aspects of control from computational to mechanical.
We encourage researchers to seek out bio-inspired techniques for co-designing morphology and control for highly integrated systems.

We also encourage considering bio-inspired principles at extreme scales.
Innovation in nano-fabrication can enable design inspired by phenomena that only occur at extremely small scales.
At the other end of the spectrum, manufacturing in microgravity outside of the Earth's atmosphere removes several constraints, potentially allowing bio-inspired design at unprecedentedly large scales.

%Bio-Exploitation
\textbf{Bio-Exploitation:}
This field has used both living and deceased biological structures, but there has been less exploration of bio-waste. 
Hair and nails have hygroscopic shape changes \cite{barba2010water}, and could be readily obtained non-destructively from a variety of mammal species. 
Many reptiles periodically shed, leaving biological tissues behind.
What other naturally expelled tissues and byproducts can be leveraged as actuators, sensors, or energy for robots?
We sincerely hope that a strong discussion of ethics and existential philosophy develops in concert with bio-exploitative robotics \cite{korsgaard2018fellow}.

%Robotic Experimental Platform
\textbf{Robotic Experimental Platform:}
We encourage researchers to leverage 3D printing and modular design strategies to rapidly probe across a parameter space \cite{urs2026robot}.
Given the ever-increasing trend in computing capability, we predict that it will be essential for robotic experimental platforms to justify their existence in comparison to simulations.
Simulations are only as good as their physics models, so robotic physical models have the advantage for tasks governed by physics phenomena that are difficult to model.
We encourage structured exploration of soft systems via combining materials with varying characteristics.

%Backspiration
\textbf{Backspiration:}
Consider using the biological counterpart as an example of the desired performance in the introduction, but reconsider using terms like ``[species]-inspired'' and ''biomimetic,'' especially in the title.
Research on convergent evolution can inspire another valid strategy: compare your non-bio-inspired designs to several biological solutions for the same problem.
Identifying common features across these systems may reveal fundamental mechanisms or constraints associated with the problem.

%%%%%%%%%%% Back matter %%%%%%%%

\subsection{Acknowledgement}
The authors would like to acknowledge current and previous students of the Evolution of Motion in Robotics and Biology Lab who helped collect references for this paper and provided helpful discussion. 
This includes, not in any particular order, 
Anvay Pradhan, 
Dr. Juri Miyamae, 
Adam Schmidt, 
John Saunders, 
Ben Andelman, 
Xiangyun Bu, 
William Olenich, 
Chae Woo Lim, 
Jongha Kim, 
Jacob Cieply, 
Dr. Xun Fu,  
Dr. Challen Enninful Adu,
Adam Hung,
Dr. Zac Brei,
Saima Jamal,
Anneliese Ferguson, 
Abdulhadi Alkayyali, 
Jessica Carlson, 
and Marco Antonio Valdez Calderon.

The authors report no competing interests.

TYM conceptualized the review. 
All authors contributed to formalizing the concepts and to writing.

MJZ is supported by a National Science Foundation Graduate Research Fellowship.

%%%%%%%%% References %%%%%%%%%%%%%%%%%

\bibliographystyle{ieeetr}
\bibliography{ref.bib}

@inproceedings{bu2025release,
  title={Release Chamber Enables Suction Cup to Delaminate and Harvest Fluid},
  author={Bu, Xiangyun and Geng, Yihao and Yin, Siyuan and Luo, Liyan and Aubin, Cameron A and Moore, Talia Y},
  booktitle={2025 IEEE 8th International Conference on Soft Robotics (RoboSoft)},
  pages={1--6},
  year={2025},
  organization={IEEE}
}

@book{korsgaard2018fellow,
  title={Fellow creatures: Our obligations to the other animals},
  author={Korsgaard, Christine Marion},
  year={2018},
  publisher={Oxford University Press}
}

@article{Cieply2026,
  title   = {Bite-Bot: A robotic platform for studying envenomation},
  author  = {Cieply, Jacob and Moore, Talia Y.},
  journal = {Integrative and Comparative Biology},
  year    = {2026},
  note    = {In revision}
}

@article{Sponberg2015InsectFlight,
  author  = {Sponberg, Simon and Dyhr, Jonathan P. and Hall, Robert W. and Daniel, Thomas L.},
  title   = {Luminance-dependent visual processing enables moth flight in low light},
  journal = {Science},
  year    = {2015},
  volume  = {348},
  number  = {6240},
  pages   = {1245--1248},
  doi     = {10.1126/science.aaa3042},
  pmid    = {26068850}
}

@article{wong2011bioinspired,
  title={Bioinspired self-repairing slippery surfaces with pressure-stable omniphobicity},
  author={Wong, Tak-Sing and Kang, Sung Hoon and Tang, Sindy KY and Smythe, Elizabeth J and Hatton, Benjamin D and Grinthal, Alison and Aizenberg, Joanna},
  journal={Nature},
  volume={477},
  number={7365},
  pages={443--447},
  year={2011},
  publisher={Nature Publishing Group UK London}
}

@article{steinhardt2021physical,
  title={A physical model of mantis shrimp for exploring the dynamics of ultrafast systems},
  author={Steinhardt, Emma and Hyun, Nak-seung P and Koh, Je-sung and Freeburn, Gregory and Rosen, Michelle H and Temel, Fatma Zeynep and Patek, SN and Wood, Robert J},
  journal={Proceedings of the National Academy of Sciences},
  volume={118},
  number={33},
  pages={e2026833118},
  year={2021},
  publisher={National Academy of Sciences}
}

@inproceedings{zhang2023launching,
  title={Launching engineered prototypes to better understand the factors that influence click beetle jump capacity},
  author={Zhang, Liyuan and Mathur, Teagan and Wissa, Aimy and Alleyne, Marianne},
  booktitle={2023 IEEE Conference on Control Technology and Applications (CCTA)},
  pages={681--686},
  year={2023},
  organization={IEEE}
}

@inproceedings{pelvic,
title = {Design and Characterisation of a Tendon-Driven
Artificial Pelvic Floor Muscle},
author = {Yael Zekaria and Antonia Tzemanaki and Jonathan Rossiter},
year = {2026},
booktitle={IEEE RoboSoft}
}

@inproceedings{berrybot,
title = {A Sensorized Blackberry Proxy for Learning Soft Fruit Harvesting},
author = {Chelse VanAtter and Srivatsan Balaji and and Joseph R. Davidson},
year = {2026},
booktitle={IEEE RoboSoft}
}

@article{kim2021biomimetic,
  title={Biomimetic chameleon soft robot with artificial crypsis and disruptive coloration skin},
  author={Kim, Hyeonseok and Choi, Joonhwa and Kim, Kyun Kyu and Won, Phillip and Hong, Sukjoon and Ko, Seung Hwan},
  journal={Nature communications},
  volume={12},
  number={1},
  pages={4658},
  year={2021},
  publisher={Nature Publishing Group UK London}
}

@article{baban2022biomimicking,
  title={Biomimicking interfacial fracture behavior of lizard tail autotomy with soft microinterlocking structures},
  author={Baban, Navajit S and Orozaliev, Ajymurat and Stubbs, Christopher J and Song, Yong-Ak},
  journal={Bioinspiration \& Biomimetics},
  volume={17},
  number={3},
  pages={036002},
  year={2022},
  publisher={IOP Publishing}
}

@article{HuaiTi,
title={TBD},
author = {Huai-Ti Lin},
journal = {Journal of Experimental Biology},
year = {2026}
}

@article{urs2026robot,
  title={The Robot Of Theseus: a modular robotic testbed for legged locomotion},
  author={Urs, Karthik and Carlson, Jessica and Manohar, Aditya Srinivas and Rakowiecki, Michael and Alkayyali, Abdulhadi and Saunders, John E and Tulbah, Faris and Moore, Talia Y},
  journal={Bioinspiration \& Biomimetics},
  volume={21},
  number={1},
  pages={016028},
  year={2026},
  publisher={IOP Publishing}
}

@article{partan2013ten,
  title={Ten unanswered questions in multimodal communication},
  author={Partan, Sarah R},
  journal={Behavioral Ecology and Sociobiology},
  volume={67},
  number={9},
  pages={1523--1539},
  year={2013}
}

@article{barba2010water,
  title={Water absorption/desorption of human hair and nails},
  author={Barba, C and Mart{\'\i}, M and Manich, AM and Carilla, J and Parra, JL and Coderch, L},
  journal={Thermochimica Acta},
  volume={503},
  pages={33--39},
  year={2010}
}

@article{ioannou2012predatory,
  title={Predatory fish select for coordinated collective motion in virtual prey},
  author={Ioannou, Christos C and Guttal, Vishwesha and Couzin, Iain D},
  journal={Science},
  volume={337},
  number={6099},
  pages={1212--1215},
  year={2012},
  publisher={American Association for the Advancement of Science}
}

@article{crowell2024ecological,
  title={Ecological drivers of ultraviolet colour evolution in snakes},
  author={Crowell, Hayley L and Curlis, John David and Weller, Hannah I and Davis Rabosky, Alison R},
  journal={Nature Communications},
  volume={15},
  number={1},
  pages={5213},
  year={2024},
  publisher={Nature Publishing Group UK London}
}

@article{li2025reverse,
  title={Reverse engineering the control law for schooling in zebrafish using virtual reality},
  author={Li, Liang and Nagy, Mate and Amichay, Guy and Wu, Ruiheng and Wang, Wei and Deussen, Oliver and Rus, Daniela and Couzin, Iain D},
  journal={Science Robotics},
  volume={10},
  number={101},
  pages={eadq6784},
  year={2025},
  publisher={American Association for the Advancement of Science}
}

@inproceedings{kohut2013precise,
  title={Precise dynamic turning of a 10 cm legged robot on a low friction surface using a tail},
  author={Kohut, Nicholas J and Pullin, Andrew O and Haldane, Duncan W and Zarrouk, David and Fearing, Ronald S},
  booktitle={2013 IEEE International Conference on Robotics and Automation},
  pages={3299--3306},
  year={2013},
  organization={IEEE}
}

@article{kim2008smooth,
  title={Smooth vertical surface climbing with directional adhesion},
  author={Kim, Sangbae and Spenko, Matthew and Trujillo, Salomon and Heyneman, Barrett and Santos, Daniel and Cutkosky, Mark R},
  journal={IEEE Transactions on robotics},
  volume={24},
  number={1},
  pages={65--74},
  year={2008},
  publisher={IEEE}
}

@article{carlsen2014bio,
  title={Bio-hybrid cell-based actuators for microsystems},
  author={Carlsen, Rika Wright and Sitti, Metin},
  journal={Small},
  volume={10},
  number={19},
  pages={3831--3851},
  year={2014},
  publisher={Wiley Online Library}
}

@article{lee2025grip,
  title={Grip and grasp: lizard claw inspired robotic manipulators},
  author={Lee, Hyeon and Douglas, Kate and Bray, Asyiah and Rummel, Andrea and Alam, Parvez},
  journal={Advanced Robotics Research},
  pages={e202500183},
  year={2025},
  publisher={Wiley Online Library}
}

@article{schultz2021climbing,
  author={Schultz, Johanna T. and Beck, Hendrik K. and Haagensen, Tina and Proost, Tasmin and Clemente, Christofer J.},
  title={Using a biologically mimicking climbing robot to explore the performance landscape of climbing in lizards},
  journal={Proceedings of the Royal Society B: Biological Sciences},
  volume={288},
  number={1947},
  pages={20202576},
  year={2021},
  doi={10.1098/rspb.2020.2576}
}

@article{li2009sensitive,
  author={Li, Chen and Umbanhowar, Paul B. and Komsuoglu, Haldun and Koditschek, Daniel E. and Goldman, Daniel I.},
  title={Sensitive dependence of the motion of a legged robot on granular media},
  journal={Proceedings of the National Academy of Sciences},
  volume={106},
  number={24},
  pages={3029--3034},
  year={2009},
  doi={10.1073/pnas.0809095106}
}

@article{saranli2001rhex,
  author={Saranli, Uluc and Buehler, Martin and Koditschek, Daniel E.},
  title={{{RHex}: A simple and highly mobile hexapod robot}},
  journal={The International Journal of Robotics Research},
  volume={20},
  number={7},
  pages={616--631},
  year={2001},
  doi={10.1177/02783640122067570}
}

@ARTICLE{niikura2022giraffe,
  author={Niikura, Atsuhiko and Nabae, Hiroyuki and Endo, Gen and Gunji, Megu and Mori, Kent and Niiyama, Ryuma and Suzumori, Koichi},
  journal={IEEE Robotics and Automation Letters}, 
  title={{Giraffe Neck Robot: First Step Toward a Powerful and Flexible Robot Prototyping Based on Giraffe Anatomy}}, 
  year={2022},
  volume={7},
  number={2},
  pages={3539-3546},
  doi={10.1109/LRA.2022.3146611}}

@book{capek1920rur,
  author = {\v{C}apek, Karel},
  title = {R.U.R. (Rossum's Universal Robots)},
  publisher = {Aventinum},
  year = {1920}
}

@inproceedings{sedal2023emergent,
  title={{Emergent Sequential Motion Through Compliant Auxetic Shells}},
  author={Sedal, Audrey and Kohler, Margaret and Agbofode, Godswill and Moore, Talia Y and Kota, Sridhar},
  booktitle={2023 IEEE/RSJ International Conference on Intelligent Robots and Systems (IROS)},
  pages={10238--10244},
  year={2023},
  organization={IEEE}
}

@inproceedings{hung2025skootr,
  title={{SKOOTR: A SKating, Omni-Oriented, Tripedal Robot}},
  author={Hung, Adam Joshua and Adu, Challen Enninful and Moore, Talia Y},
  booktitle={2025 IEEE International Conference on Robotics and Automation (ICRA)},
  pages={15921--15928},
  year={2025},
  organization={IEEE}
}

@article{hannaford1995anthroform,
  title={The anthroform biorobotic arm: a system for the study of spinal circuits},
  author={Hannaford, Blake and Winters, Jack M and Chou, Ching-Ping and Marbot, Pierre-Henry},
  journal={Annals of biomedical engineering},
  volume={23},
  number={4},
  pages={399--408},
  year={1995},
  publisher={Springer}
}

@article{ramezani2017biomimetic,
  title={A biomimetic robotic platform to study flight specializations of bats},
  author={Ramezani, Alireza and Chung, Soon-Jo and Hutchinson, Seth},
  journal={Science Robotics},
  volume={2},
  number={3},
  pages={eaal2505},
  year={2017},
  publisher={American Association for the Advancement of Science}
}

@article{nyakatura2019reverse,
  title={Reverse-engineering the locomotion of a stem amniote},
  author={Nyakatura, John A and Melo, Kamilo and Horvat, Tomislav and Karakasiliotis, Kostas and Allen, Vivian R and Andikfar, Amir and Andrada, Emanuel and Arnold, Patrick and Laustr{\"o}er, Jonas and Hutchinson, John R and others},
  journal={Nature},
  volume={565},
  number={7739},
  pages={351--355},
  year={2019},
  publisher={Nature Publishing Group UK London}
}

@article{harvey2026bio,
  title={{How bio-inspired is your design? A transparent reporting framework}},
  author={Harvey, Christina},
  journal={Communications Engineering},
  volume={5},
  number={1},
  pages={70},
  year={2026},
  publisher={Nature Publishing Group UK London}
}

@article{kim2026dead,
  title={{Dead Matter, Living Machines: Repurposing Crustaceans' Abdomen Exoskeleton for Bio-Hybrid Robots}},
  author={Kim, Sareum and Gilday, Kieran and Hughes, Josie},
  journal={Advanced Science},
  volume={13},
  number={15},
  pages={e17712},
  year={2026},
  publisher={Wiley Online Library}
}

@article{kriegman2020scalable,
  title={A scalable pipeline for designing reconfigurable organisms},
  author={Kriegman, Sam and Blackiston, Douglas and Levin, Michael and Bongard, Josh},
  journal={Proceedings of the National Academy of Sciences},
  volume={117},
  number={4},
  pages={1853--1859},
  year={2020},
  publisher={National Academy of Sciences}
}

@article{jayaram2016cockroaches,
  title={Cockroaches traverse crevices, crawl rapidly in confined spaces, and inspire a soft, legged robot},
  author={Jayaram, Kaushik and Full, Robert J},
  journal={Proceedings of the National Academy of Sciences},
  volume={113},
  number={8},
  pages={E950--E957},
  year={2016},
  publisher={National Academy of Sciences}
}

@misc{spot,
title={Spot},
author={Boston Dynamics},
url={https://www.bostondynamics.com/products/spot }
}

@article{haldane2016robotic,
  title={Robotic vertical jumping agility via series-elastic power modulation},
  author={Haldane, Duncan W and Plecnik, Mark M and Yim, Justin K and Fearing, Ronald S},
  journal={Science Robotics},
  volume={1},
  number={1},
  pages={eaag2048},
  year={2016},
  publisher={American Association for the Advancement of Science}
}

@article{ijspeert2007swimming,
  title={From swimming to walking with a salamander robot driven by a spinal cord model},
  author={Ijspeert, Auke Jan and Crespi, Alessandro and Ryczko, Dimitri and Cabelguen, Jean-Marie},
  journal={science},
  volume={315},
  number={5817},
  pages={1416--1420},
  year={2007},
  publisher={American Association for the Advancement of Science}
}

@article{yoo2023compliant,
  title={Compliant suction gripper with seamless deployment and retraction for robust picking against depth and tilt errors},
  author={Yoo, Yuna and Eom, Jaemin and Park, MinJo and Cho, Kyu-Jin},
  journal={IEEE Robotics and Automation Letters},
  volume={8},
  number={3},
  pages={1311--1318},
  year={2023},
  publisher={IEEE}
}

@article{Libby2012,
author = {Libby, Thomas and Moore, Talia Yuki and Chang-Siu, Evan and Li, Deborah and Cohen, Daniel J. and Jusufi, Ardian and Full, Robert J.},
doi = {10.1038/nature10710},
journal = {Nature},
month = {jan},
number = {7380},
pages = {181--184},
pmid = {22217942},
publisher = {Nature Publishing Group},
title = {{Tail-assisted pitch control in lizards, robots and dinosaurs}},
volume = {481},
year = {2012}
}

@inproceedings{danforth2020emulating,
  title={Emulating duration and curvature of coral snake anti-predator thrashing behaviors using a soft-robotic platform},
  author={Danforth, Shannon M and Kohler, Margaret and Bruder, Daniel and Rabosky, Alison R Davis and Kota, Sridhar and Vasudevan, Ram and Moore, Talia Y},
  booktitle={2020 IEEE International Conference on Robotics and Automation (ICRA)},
  pages={5068--5074},
  year={2020},
  organization={IEEE}
}

@inproceedings{hutter2016anymal,
  title={{ANYmal - a highly mobile and dynamic quadrupedal robot}},   
  author={Hutter, Marco and Gehring, Christian and Jud, Dominic and Lauber, Andreas and Bellicoso, C. Dario and Tsounis, Vassilios and Hwangbo, Jemin and Bodie, Karen and Fankhauser, Peter and Bloesch, Michael and Diethelm, Remo and Bachmann, Samuel and Melzer, Amir and Hoepflinger, Mark},
  booktitle={2016 IEEE/RSJ International Conference on Intelligent Robots and Systems (IROS)}, 
  year={2016},
  pages={38-44},
  doi={10.1109/IROS.2016.7758092}
}

@inproceedings{bingham2024biomimetic,
  title={Biomimetic Design and Development of an Oryctodromeus-Inspired Robotic Dinosaur Skeleton},
  author={Bingham, Kyler and Hafezi, Amir and Thapa, Anish and Sourani Yancheshmeh, Sara and Zakrevski, Christopher and Berry, Matthew and Das, Shaibal and Walker, Payton and Cortez Lopez, Juan and Thompson, Dominik and others},
  booktitle={ASME International Mechanical Engineering Congress and Exposition},
  volume={88636},
  pages={V005T07A028},
  year={2024},
  organization={American Society of Mechanical Engineers}
}

@article{dicks_philosophy_2016,
	title = {{The Philosophy of Biomimicry}},
	volume = {29},
	doi = {10.1007/s13347-015-0210-2},
	pages = {223--243},
	number = {3},
	journal = {Philosophy \& Technology},
	author = {Dicks, Henry},
	year = {2016}
}

@article{vickers2000mechanisms,
  title={Mechanisms of animal navigation in odor plumes},
  author={Vickers, Neil J},
  journal={The Biological Bulletin},
  volume={198},
  number={2},
  pages={203--212},
  year={2000},
  publisher={Marine Biological Laboratory}
}

@incollection{barth1998vibrational,
  author    = {Barth, Friedrich G.},
  title     = {The Vibrational Sense of Spiders},
  booktitle = {Comparative Hearing: Insects},
  editor    = {Hoy, Ronald R. and Popper, Arthur N. and Fay, Richard R.},
  series    = {Springer Handbook of Auditory Research},
  volume    = {10},
  pages     = {228--278},
  publisher = {Springer},
  address   = {New York, NY},
  year      = {1998},
  doi       = {10.1007/978-1-4612-0585-2_7}
}

@book{habib_bioinspiration_2007,
	location = {Erscheinungsort nicht ermittelbar},
	title = {{Bioinspiration and Robotics: Walking and Climbing Robots}},
	isbn = {978-3-902613-15-8 978-953-51-5814-1},
	publisher = {{IntechOpen}},
	editor = {Habib, Maki},
	year = {2007}
}

@incollection{iida_biologically_2016,
	location = {Cham},
	title = {{Biologically Inspired Robotics}},
	isbn = {978-3-319-32552-1},
	pages = {2015--2034},
	booktitle = {Springer Handbook of Robotics},
	publisher = {Springer International Publishing},
	author = {Iida, Fumiya and Ijspeert, Auke Jan},
	editor = {Siciliano, Bruno and Khatib, Oussama},
	year = {2016},
	doi = {10.1007/978-3-319-32552-1_75}
}

@article{ramdya2023neuromechanics,
  title={{The neuromechanics of animal locomotion: From biology to robotics and back}},
  author={Ramdya, Pavan and Ijspeert, Auke Jan},
  journal={Science Robotics},
  volume={8},
  number={78},
  pages={eadg0279},
  year={2023},
  publisher={American Association for the Advancement of Science}
}

@article{wanieck_perception_2021,
	title = {Perception and role of standards in the world of biomimetics},
	volume = {10},
    year = {2021},
	doi = {10.1680/jbibn.20.00024},
	pages = {8--15},
	number = {1},
	journal = {Bioinspired, Biomimetic and Nanobiomaterials},
	author = {Wanieck, Kristina and Beismann, Heike}
}

@inproceedings{sharma_biomimicry_2019,
    year = {2019},
	title = {{Biomimicry: Exploring Research, Challenges, Gaps, and Tools}},
	doi = {10.1007/978-981-13-5974-3_8},
	pages = {87--97},
	booktitle = {Research into Design for a Connected World},
	publisher = {Springer},
	author = {Sharma, Sunil and Sarkar, Prabir},
	editor = {Chakrabarti, Amaresh}
}

@article{fu_bio-inspired_2014,
	title = {{Bio-Inspired Design: An Overview Investigating Open Questions From the Broader Field of Design-by-Analogy}},
	volume = {136},
	doi = {10.1115/1.4028289},
	number = {111102},
	journal = {Journal of Mechanical Design},
	author = {Fu, Katherine and Moreno, Diana and Yang, Maria and Wood, Kristin L.},
	year = {2014}
}

@incollection{sansoni2016aesthetic,
  title={Aesthetic of prosthetic devices: from medical equipment to a work of design},
  author={Sansoni, Stefania and Speer, Leslie and Wodehouse, Andrew and Buis, Arjan},
  booktitle={Emotional Engineering Volume 4},
  pages={73--92},
  year={2016},
  publisher={Springer}
}

@article{razoki2025biomimetic,
  title={Biomimetic Strategies in Kinetic Architecture: A Comparative Analysis of Nature-Inspired Roof and Fa{\c{c}}ade Designs},
  author={Razoki, F and Al-Kazzaz, D},
  journal={International Journal of Design and Nature and Ecodynamics},
  volume={20},
  pages={1269--1282},
  year={2025}
}

@inproceedings{koditschek_principled_2004,
	title = {A principled approach to bio-inspired design of legged locomotion systems},
	volume = {5422},
	doi = {10.1117/12.544118},
	pages = {86--100},
	booktitle = {Unmanned Ground Vehicle Technology {VI}},
	publisher = {{SPIE}},
	author = {Koditschek, Daniel E. and Full, Robert J. and Buehler, Martin},
	year = {2004},
}

@article{barley_addressing_2022,
	title = {{Addressing Diverse Motivations to Enable Bioinspired Design}},
	volume = {62},
	doi = {10.1093/icb/icac041},
	number = {5},
	journal = {Integrative and Comparative Biology},
	author = {Barley, William C and Ruge-Jones, Luisa and Wissa, Aimy and Suarez, Andrew V and Alleyne, Marianne},
	year = {2022}
}

@inproceedings{lynch2022autonomous,
  title={Autonomous actuation of flapping wing robots inspired by asynchronous insect muscle},
  author={Lynch, James and Gau, Jeff and Sponberg, Simon and Gravish, Nick},
  booktitle={2022 International Conference on Robotics and Automation (ICRA)},
  pages={2076--2083},
  year={2022},
  organization={IEEE}
}

@article{gravish_robotics-inspired_2018,
	title = {{Robotics-inspired biology}},
	volume = {221},
	doi = {10.1242/jeb.138438},
	number = {7},
	journal = {Journal of Experimental Biology},
	author = {Gravish, Nick and Lauder, George V.},
	year = {2018}
}

@article{lauder_robotics_2022,
	title = {{Robotics as a Comparative Method in Ecology and Evolutionary Biology}},
	volume = {62},
	doi = {10.1093/icb/icac016},
	pages = {721--734},
	number = {3},
	journal = {Integrative and Comparative Biology},
	author = {Lauder, George V},
	year = {2022}
}

@article{white_tunabot_2021,
	title = {{Tunabot Flex: a tuna-inspired robot with body flexibility improves high-performance swimming}},
	volume = {16},
	doi = {10.1088/1748-3190/abb86d},
	pages = {026019},
	number = {2},
	journal = {Bioinspiration \& Biomimetics},
	author = {White, Carl H and Lauder, George V and Bart-Smith, Hilary},
	year = {2021}
}

@article{roberts_testing_2014,
	title = {{Testing Biological Hypotheses with Embodied Robots: Adaptations, Accidents, and By-Products in the Evolution of Vertebrates}},
	volume = {1},
	doi = {10.3389/frobt.2014.00012},
	journal= {Frontiers in Robotics and {AI}},
	author = {Roberts, Sonia F. and Hirokawa, Jonathan and Rosenblum, Hannah G. and Sakhtah, Hassan and Gutierrez, Andres A. and Porter, Marianne E. and Long, John H.},
	year = {2014}
}

@article{flammang_bioinspired_2022,
	title = {{Bioinspired Design in Research: Evolution as Beta-Testing}},
	volume = {62},
	doi = {10.1093/icb/icac134},
	pages = {1164--1173},
	number = {5},
	journal = {Integrative and Comparative Biology},
	author = {Flammang, Brooke E},
	year = {2022}
}

@article{holmes_dynamics_2006,
	title = {{The Dynamics of Legged Locomotion: Models, Analyses, and Challenges}},
	volume = {48},
	doi = {10.1137/S0036144504445133},
	pages = {207--304},
	number = {2},
	journal = {{SIAM} Review},
	author = {Holmes, Philip and Full, Robert J. and Koditschek, Dan and Guckenheimer, John},
	year = {2006}
}

@article{tamborini_is_2022,
	title = {{Is biorobotics science? Some theoretical reflections}},
	volume = {18},
	doi = {10.1088/1748-3190/aca24b},
	pages = {015005},
	number = {1},
	journal= {Bioinspiration \& Biomimetics},
	author = {Tamborini, Marco and Datteri, Edoardo},
	year = {2022}
}

@book{webb_biorobotics_2001,
	location = {Cambridge, {MA}, {USA}},
	title = {Biorobotics},
	isbn = {978-0-262-73141-6},
	publisher = {{MIT} Press},
	author = {Webb, Barbara and Consilvio, Thomas},
	year = {2001-07},
}

@article{raman_modular_2017,
	title = {A modular approach to the design, fabrication, and characterization of muscle-powered biological machines},
	volume = {12},
	doi = {10.1038/nprot.2016.185},
	pages = {519--533},
	number = {3},
	journal = {Nature Protocols},
	author = {Raman, Ritu and Cvetkovic, Caroline and Bashir, Rashid},
	year = {2017}
}

@article{cvetkovic_three-dimensionally_2014,
	title = {Three-dimensionally printed biological machines powered by skeletal muscle},
	volume = {111},
	doi = {10.1073/pnas.1401577111},
	pages = {10125--10130},
	number = {28},
	journal = {Proceedings of the National Academy of Sciences},
	author = {Cvetkovic, Caroline and Raman, Ritu and Chan, Vincent and Williams, Brian J. and Tolish, Madeline and Bajaj, Piyush and Sakar, Mahmut Selman and Asada, H. Harry and Saif, M. Taher A. and Bashir, Rashid},
	year = {2014}
}

@article{raman_biofabrication_2024,
	title = {{Biofabrication of Living Actuators}},
	volume = {26},
	doi = {10.1146/annurev-bioeng-110122-013805},
	pages = {223--245},
	issue = {Volume 26, 2024},
	journal = {Annual Review of Biomedical Engineering},
	author = {Raman, Ritu},
	year = {2024}
}

@article{dogan_immune_2024,
	title = {{Immune Cell-Based Microrobots for Remote Magnetic Actuation, Antitumor Activity, and Medical Imaging}},
	volume = {13},
	doi = {10.1002/adhm.202400711},
	pages = {2400711},
	number = {23},
	journal = {Advanced Healthcare Materials},
	author = {Dogan, Nihal Olcay and Suadiye, Eylül and Wrede, Paul and Lazovic, Jelena and Dayan, Cem Balda and Soon, Ren Hao and Aghakhani, Amirreza and Richter, Gunther and Sitti, Metin},
	year = {2024}
}

@article{filippi_multicellular_2025,
	title = {Multicellular muscle-tendon bioprinting of mechanically optimized musculoskeletal bioactuators with enhanced force transmission},
	volume = {11},
	doi = {10.1126/sciadv.adv2628},
	pages = {eadv2628},
	number = {29},
	journal = {Science Advances},
	author = {Filippi, Miriam and Mock, Diana and Fuentes, Judith and Y. Michelis, Mike and Balciunaite, Aiste and Paniagua, Pablo and Hopf, Raoul and Barteld, Adina and Eng, Selina and Badolato, Asia and Snedeker, Jess and Guix, Maria and Sanchez, Samuel and K. Katzschmann, Robert},
	year = {2025}
}

@inproceedings{sato_cyborg_2008,
	title = {{A cyborg beetle: Insect flight control through an implantable, tetherless microsystem}},
	doi = {10.1109/MEMSYS.2008.4443618},
	pages = {164--167},
	booktitle = {2008 {IEEE} 21st International Conference on Micro Electro Mechanical Systems},
	author = {Sato, Hirotaka and Berry, Christopher W. and Casey, Brendan E. and Lavella, Gabriel and Yao, Ying and {VandenBrooks}, John M. and Maharbiz, Michel M.},
	year = {2008}
}

@article{burden_why_2024,
	title = {Why animals can outrun robots},
	volume = {9},
	doi = {10.1126/scirobotics.adi9754},
	pages = {eadi9754},
	number = {89},
	journal = {Science Robotics},
	author = {Burden, Samuel A. and Libby, Thomas and Jayaram, Kaushik and Sponberg, Simon and Donelan, J. Maxwell},
	year = {2024}
}

@article{mishra_sensorimotor_2024,
	title = {Sensorimotor control of robots mediated by electrophysiological measurements of fungal mycelia},
	volume = {9},
	doi = {10.1126/scirobotics.adk8019},
	pages = {eadk8019},
	number = {93},
	journal = {Science Robotics},
	author = {Mishra, Anand Kumar and Kim, Jaeseok and Baghdadi, Hannah and Johnson, Bruce R. and Hodge, Kathie T. and Shepherd, Robert F.},
	year = {2024}
}

@article{yap_necrobotics_2022,
	title = {{Necrobotics: Biotic Materials as Ready-to-Use Actuators}},
	volume = {9},
	doi = {10.1002/advs.202201174},
	pages = {2201174},
	number = {29},
	journal = {Advanced Science},
	author = {Yap, Te Faye and Liu, Zhen and Rajappan, Anoop and Shimokusu, Trevor J. and Preston, Daniel J.},
	year = {2022}
}

@article{puma_3d_2025,
	title = {{3D necroprinting: Leveraging biotic material as the nozzle for 3D printing}},
	volume = {11},
	doi = {10.1126/sciadv.adw9953},
	pages = {eadw9953},
	number = {47},
	journal = {Science Advances},
	author = {Puma, Justin and Yang, Zhen and Johnston, Evan and Zhang, Zixin and Lan, Xiaoyi and Zhang, Lingzhi and Hou, Hongyu and He, Zixin and Afify, Ali and Creighton, Megan A. and Li, Jianyu and Cao, Changhong},
	year = {2025-11-19}
}

@article{agerholm1961artificial,
  title={The" artificial muscle" of McKibben},
  author={Agerholm, Margaret and Lord, Alphonsus},
  journal={The Lancet},
  volume={277},
  number={7178},
  pages={660--661},
  year={1961},
  publisher={Elsevier}
}

@article{kim_dead_2026,
	title = {{Dead Matter, Living Machines: Repurposing Crustaceans' Abdomen Exoskeleton for Bio-Hybrid Robots}},
	volume = {13},
	doi = {10.1002/advs.202517712},
	pages = {e17712},
	number = {15},
	journal = {Advanced Science},
	author = {Kim, Sareum and Gilday, Kieran and Hughes, Josie},
	year = {2026}
}

@article{wang2025spirobs,
  title={SpiRobs: Logarithmic spiral-shaped robots for versatile grasping across scales},
  author={Wang, Zhanchi and Freris, Nikolaos M and Wei, Xi},
  journal={Device},
  volume={3},
  number={4},
  year={2025},
  publisher={Elsevier}
}

@article{libby_tail-assisted_2012,
	title = {Tail-assisted pitch control in lizards, robots and dinosaurs},
	volume = {481},
	doi = {10.1038/nature10710},
	pages = {181--184},
	number = {7380},
	journal = {Nature},
	author = {Libby, Thomas and Moore, Talia Y. and Chang-Siu, Evan and Li, Deborah and Cohen, Daniel J. and Jusufi, Ardian and Full, Robert J.},
	year = {2012},
}

@inproceedings{azocar2018design,
  title={Design and characterization of an open-source robotic leg prosthesis},
  author={Azocar, Alejandro F and Mooney, Luke M and Hargrove, Levi J and Rouse, Elliott J},
  booktitle={2018 7th IEEE International Conference on Biomedical Robotics and Biomechatronics (Biorob)},
  pages={111--118},
  year={2018},
  organization={IEEE}
}

@article{patricelli2010tactical,
  title={Tactical allocation of effort among multiple signals in sage grouse: an experiment with a robotic female},
  author={Patricelli, Gail L and Krakauer, Alan H},
  journal={Behavioral Ecology},
  volume={21},
  number={1},
  pages={97--106},
  year={2010},
  publisher={Oxford University Press}
}

@inproceedings{asbeck2006climbing,
  title={Climbing walls with microspines},
  author={Asbeck, Alan T and Kim, Sangbae and McClung, Arthur and Parness, Aaron and Cutkosky, Mark R},
  booktitle={IEEE ICRA},
  pages={4315--4317},
  year={2006},
  organization={Fla.}
}

@inbook{chen_feed_2023,
	location = {Cham},
	title = {{Feed Me: Robotic Infiltration of Poison Frog Families}},
	isbn = {978-3-031-39504-8},
	doi = {10.1007/978-3-031-39504-8_20},
	pages = {293--302},
	booktitle = {Biomimetic and Biohybrid Systems},
	publisher = {Springer Nature Switzerland},
	author = {Chen, Tony G. and Goolsby, Billie C. and Bernal, Guadalupe and O’Connell, Lauren A. and Cutkosky, Mark R.},
	year = {2023}
}

@article{dufour_recent_2020,
	title = {{Recent biological invasion shapes species recognition and aggressive behaviour in a native species: A behavioural experiment using robots in the field}},
	volume = {89},
	doi = {10.1111/1365-2656.13223},
	pages = {1604--1614},
	number = {7},
	journal = {Journal of Animal Ecology},
	author = {Dufour, Claire M. S. and Clark, David L. and Herrel, Anthony and Losos, Jonathan B.},
	year = {2020}
}

@book{carver_mead_book,
author = {Mead, Carver},
title = {Analog VLSI and neural systems},
year = {1989},
isbn = {0201059924},
publisher = {Addison-Wesley Longman Publishing Co., Inc.},
address = {USA}}

@ARTICLE{Loihi,
  author={Davies, Mike and Srinivasa, Narayan and Lin, Tsung-Han and Chinya, Gautham and Cao, Yongqiang and Choday, Sri Harsha and Dimou, Georgios and Joshi, Prasad and Imam, Nabil and Jain, Shweta and Liao, Yuyun and Lin, Chit-Kwan and Lines, Andrew and Liu, Ruokun and Mathaikutty, Deepak and McCoy, Steven and Paul, Arnab and Tse, Jonathan and Venkataramanan, Guruguhanathan and Weng, Yi-Hsin and Wild, Andreas and Yang, Yoonseok and Wang, Hong},
  journal={IEEE Micro}, 
  title={Loihi: A Neuromorphic Manycore Processor with On-Chip Learning}, 
  year={2018},
  volume={38},
  number={1},
  pages={82-99},
  doi={10.1109/MM.2018.112130359}}

@ARTICLE{BrainScale2,
AUTHOR={Pehle, Christian  and Billaudelle, Sebastian  and Cramer, Benjamin  and Kaiser, Jakob  and Schreiber, Korbinian  and Stradmann, Yannik  and Weis, Johannes  and Leibfried, Aron  and Müller, Eric  and Schemmel, Johannes },      
TITLE={The BrainScaleS-2 Accelerated Neuromorphic System With Hybrid Plasticity},    
JOURNAL={Frontiers in Neuroscience},     
VOLUME={Volume 16 - 2022},
YEAR={2022},
DOI={10.3389/fnins.2022.795876},
ISSN={1662-453X},
}

@ARTICLE{TrueNorth,
  author={Akopyan, Filipp and Sawada, Jun and Cassidy, Andrew and Alvarez-Icaza, Rodrigo and Arthur, John and Merolla, Paul and Imam, Nabil and Nakamura, Yutaka and Datta, Pallab and Nam, Gi-Joon and Taba, Brian and Beakes, Michael and Brezzo, Bernard and Kuang, Jente B. and Manohar, Rajit and Risk, William P. and Jackson, Bryan and Modha, Dharmendra S.},
  journal={IEEE Transactions on Computer-Aided Design of Integrated Circuits and Systems}, 
  title={TrueNorth: Design and Tool Flow of a 65 mW 1 Million Neuron Programmable Neurosynaptic Chip}, 
  year={2015},
  volume={34},
  number={10},
  pages={1537-1557},
  doi={10.1109/TCAD.2015.2474396}}

@misc{SpiNNaker2,
      title={SpiNNaker2: A Large-Scale Neuromorphic System for Event-Based and Asynchronous Machine Learning}, 
      author={Hector A. Gonzalez and Jiaxin Huang and Florian Kelber and Khaleelulla Khan Nazeer and Tim Langer and Chen Liu and Matthias Lohrmann and Amirhossein Rostami and Mark Schöne and Bernhard Vogginger and Timo C. Wunderlich and Yexin Yan and Mahmoud Akl and Christian Mayr},
      year={2024},
      eprint={2401.04491},
      archivePrefix={arXiv},
      primaryClass={cs.ET},
      url={https://arxiv.org/abs/2401.04491}, 
}

@article{NeuromorphicSiliconNeuron2011,
  title = {Neuromorphic {{Silicon Neuron Circuits}}},
  author = {Indiveri, Giacomo and {Linares-Barranco}, Bernabe and Hamilton, Tara Julia and {van Schaik}, Andr{\'e} and {Etienne-Cummings}, Ralph and Delbruck, Tobi and Liu, Shih-Chii and Dudek, Piotr and H{\"a}fliger, Philipp and Renaud, Sylvie and Schemmel, Johannes and Cauwenberghs, Gert and Arthur, John and Hynna, Kai and Folowosele, Fopefolu and Sa{\"i}ghi, Sylvain and {Serrano-Gotarredona}, Teresa and Wijekoon, Jayawan and Wang, Yingxue and Boahen, Kwabena},
  year = 2011,
  journal = {Frontiers in Neuroscience},
  volume = {5},
  publisher = {Frontiers},
  doi = {10.3389/fnins.2011.00073},
}

@Article{neuro_photonics,
AUTHOR = {Kutluyarov, Ruslan V. and Zakoyan, Aida G. and Voronkov, Grigory S. and Grakhova, Elizaveta P. and Butt, Muhammad A.},
TITLE = {Neuromorphic Photonics Circuits: Contemporary Review},
JOURNAL = {Nanomaterials},
VOLUME = {13},
YEAR = {2023},
NUMBER = {24},
ARTICLE-NUMBER = {3139},
PubMedID = {38133036},
DOI = {10.3390/nano13243139}
}

@ARTICLE{neuro_magnets,
AUTHOR={Wang, Cheng  and Agrawal, Amogh  and Yu, Eunseon  and Roy, Kaushik },     
TITLE={Multi-Level Neuromorphic Devices Built on Emerging Ferroic Materials: A Review},     
JOURNAL={Frontiers in Neuroscience},    
VOLUME={Volume 15 - 2021},
YEAR={2021},
DOI={10.3389/fnins.2021.661667},
ISSN={1662-453X},
}

@article{neuro_memristors,
    author = {Kim, So-Yeon and Zhang, Heyi and Rivera-Sierra, Gonzalo and Fenollosa, Roberto and Rubio-Magnieto, Jenifer and Bisquert, Juan},
    title = {Introduction to neuromorphic functions of memristors: The inductive nature of synapse potentiation},
    journal = {Journal of Applied Physics},
    volume = {137},
    number = {11},
    pages = {111101},
    year = {2025},
    month = {03},
    doi = {10.1063/5.0257462}
}

@article{hooker2020hardwarelottery,
  title={The hardware lottery},
  author={Hooker, Sara},
  journal={Communications of the ACM},
  volume={64},
  number={12},
  pages={58--65},
  year={2021},
  publisher={ACM New York, NY, USA}
}

@article{Event_Based_Survey,
author={Gallego, Guillermo and Delbruck, Tobi and Orchard, Garrick and Bartolozzi, Chiara and Taba, Brian and Censi, Andrea and Leutenegger, Stefan and Davison, Andrew J. and Conradt, Jorg and Daniilidis, Kostas and Scaramuzza, Davide},
journal={ IEEE Transactions on Pattern Analysis \& Machine Intelligence },
title={{ Event-Based Vision: A Survey }},
year={2022},
volume={44},
number={01},
ISSN={1939-3539},
pages={154-180},
doi={10.1109/TPAMI.2020.3008413},
publisher={IEEE Computer Society},
address={Los Alamitos, CA, USA}}

@INPROCEEDINGS{Event_Control_Quad,
  author={Dimitrova, Rika Sugimoto and Gehrig, Mathias and Brescianini, Dario and Scaramuzza, Davide},
  booktitle={2020 IEEE International Conference on Robotics and Automation (ICRA)}, 
  title={Towards Low-Latency High-Bandwidth Control of Quadrotors using Event Cameras}, 
  year={2020},
  volume={},
  number={},
  pages={4294-4300},
  doi={10.1109/ICRA40945.2020.9197530}}

@INPROCEEDINGS{Event_VIO,
  author={Zhu, Alex Zihao and Atanasov, Nikolay and Daniilidis, Kostas},
  booktitle={2017 IEEE Conference on Computer Vision and Pattern Recognition (CVPR)}, 
  title={Event-Based Visual Inertial Odometry}, 
  year={2017},
  volume={},
  number={},
  pages={5816-5824},
  doi={10.1109/CVPR.2017.616}}

@article{event_obstacle_avoidance,
author = {Davide Falanga  and Kevin Kleber  and Davide Scaramuzza },
title = {Dynamic obstacle avoidance for quadrotors with event cameras},
journal = {Science Robotics},
volume = {5},
number = {40},
pages = {eaaz9712},
year = {2020},
doi = {10.1126/scirobotics.aaz9712}
}

@Inbook{Mahowald1994,
author="Mahowald, Misha",
title="The Silicon Retina",
bookTitle="An Analog VLSI System for Stereoscopic Vision",
year="1994",
publisher="Springer US",
address="Boston, MA",
pages="4--65",
isbn="978-1-4615-2724-4",
doi="10.1007/978-1-4615-2724-4_2"
}

@ARTICLE{nengo, 
AUTHOR={DeWolf, Travis  and Jaworski, Pawel  and Eliasmith, Chris },      
TITLE={Nengo and Low-Power AI Hardware for Robust, Embedded Neurorobotics},        
JOURNAL={Frontiers in Neurorobotics},        
VOLUME={Volume 14 - 2020},
YEAR={2020},
DOI={10.3389/fnbot.2020.568359},
ISSN={1662-5218}
}

@inproceedings{neuromorphic_PID,
  author    = {Rasmus Stagsted and Antonio Vitale and Jonas Binz and Alpha Renner and Leon Bonde Larsen and Yulia Sandamirskaya},
  title     = {{Towards neuromorphic control: A spiking neural network based PID controller for UAV}},
  booktitle = {Proceedings of Robotics: Science and Systems (RSS)},
  year      = {2020},
  doi       = {10.15607/RSS.2020.XVI.074}
  }

@article{neuromorphic_MPC,
  author  = {Ashish Rao Mangalore and Gabriel Andr{\'e}s Fonseca Guerra and Sumedh R. Risbud and Philipp Stratmann and Andreas Wild},
  title   = {Neuromorphic Quadratic Programming for Efficient and Scalable Model Predictive Control: Towards Advancing Speed and Energy Efficiency in Robotic Control},
  journal = {IEEE Robotics \& Automation Magazine},
  year    = {2024},
  volume  = {32},
  number  = {2},
  pages   = {69--79},
  doi     = {10.1109/MRA.2024.3415005}
}

@INPROCEEDINGS{SNN_SLAM,
  author={Tang, Guangzhi and Shah, Arpit and Michmizos, Konstantinos P.},
  booktitle={2019 IEEE/RSJ International Conference on Intelligent Robots and Systems (IROS)}, 
  title={Spiking Neural Network on Neuromorphic Hardware for Energy-Efficient Unidimensional SLAM}, 
  year={2019},
  volume={},
  number={},
  pages={4176-4181},
  doi={10.1109/IROS40897.2019.8967864}}

@article{Zanatta2024ExploringSNNDRLRobotics,
  author  = {Zanatta, Luca and Barchi, Francesco and Manoni, Simone and Tolu, Silvia and Bartolini, Andrea and Acquaviva, Andrea},
  title   = {Exploring spiking neural networks for deep reinforcement learning in robotic tasks},
  journal = {Scientific Reports},
  year    = {2024},
  volume  = {14},
  number  = {1},
  pages   = {30648},
  doi     = {10.1038/s41598-024-77779-8},
  url     = {https://www.nature.com/articles/s41598-024-77779-8}
}

@article{why_systolic,
author = {Kung, H. T.},
title = {Why Systolic Architectures?},
year = {1982},
issue_year = {January 1982},
publisher = {IEEE Computer Society Press},
address = {Washington, DC, USA},
volume = {15},
number = {1},
doi = {10.1109/MC.1982.1653825},
journal = {Computer},
pages = {37–46},
numpages = {10}
}

@ARTICLE{PIM,
  author={Ghose, S. and Boroumand, A. and Kim, J. S. and Gómez-Luna, J. and Mutlu, O.},
  journal={IBM Journal of Research and Development}, 
  title={Processing-in-memory: A workload-driven perspective}, 
  year={2019},
  volume={63},
  number={6},
  pages={3:1-3:19},
  doi={10.1147/JRD.2019.2934048}}

@article{Poggio2012LevelsOfUnderstanding,
  author  = {Poggio, Tomaso A.},
  title   = {The Levels of Understanding Framework, Revised},
  journal = {Perception},
  year    = {2012},
  volume  = {41},
  number  = {9},
  pages   = {1017--1023},
  doi     = {10.1068/p7299}
}

@article {digital_sphinx,
	author = {Brunton, Bingni W. and Abe, Elliott T.T. and Hu, Lawrence Jianqiao and Tuthill, John C.},
	title = {The digital sphinx: Can a worm brain control a fly body?},
	year = {2026},
	doi = {10.64898/2026.03.20.713233},
	publisher = {Cold Spring Harbor Laboratory},
	journal = {bioRxiv}
}

@article{neuroscience_microprocessor,
  title={Could a neuroscientist understand a microprocessor?},
  author={Jonas, Eric and Kording, Konrad Paul},
  journal={PLoS computational biology},
  volume={13},
  number={1},
  pages={e1005268},
  year={2017},
  publisher={Public Library of Science}
}

@article{MetaphorBrainsAsComputers,
  author  = {Brette, Romain},
  title   = {Brains as Computers: Metaphor, Analogy, Theory or Fact?},
  journal = {Frontiers in Ecology and Evolution},
  year    = {2022},
  volume  = {10},
  pages   = {878729},
  doi     = {10.3389/fevo.2022.878729}
}

@article{WhenDoesPhysicalSystemCompute,
  author  = {Horsman, Dominic and Stepney, Susan and Wagner, Rob C. and Kendon, Viv},
  title   = {When does a physical system compute?},
  journal = {Proceedings of the Royal Society A: Mathematical, Physical and Engineering Sciences},
  year    = {2014},
  volume  = {470},
  number  = {2169},
  pages   = {20140182},
  doi     = {10.1098/rspa.2014.0182}
}

@ARTICLE{Loihi_compiler,
  author={Lin, Chit-Kwan and Wild, Andreas and Chinya, Gautham N. and Cao, Yongqiang and Davies, Mike and Lavery, Daniel M. and Wang, Hong},
  journal={Computer}, 
  title={Programming Spiking Neural Networks on Intel’s Loihi}, 
  year={2018},
  volume={51},
  number={3},
  pages={52-61},
  doi={10.1109/MC.2018.157113521}}

@article{Boahen2000AERPointToPoint,
  author  = {Boahen, Kwabena A.},
  title   = {Point-to-Point Connectivity Between Neuromorphic Chips Using Address Events},
  journal = {IEEE Transactions on Circuits and Systems II: Analog and Digital Signal Processing},
  year    = {2000},
  volume  = {47},
  number  = {5},
  pages   = {416--434},
  doi     = {10.1109/82.842110}
}

@article{NeuroInspiredAI,
  author  = {Hassabis, Demis and Kumaran, Dharshan and Summerfield, Christopher and Botvinick, Matthew},
  title   = {Neuroscience-Inspired Artificial Intelligence},
  journal = {Neuron},
  year    = {2017},
  volume  = {95},
  number  = {2},
  pages   = {245--258},
  doi     = {10.1016/j.neuron.2017.06.011}
}

@article{evolutionary_algorithms,
  author  = {Hua, Yicun and Liu, Qiqi and Hao, Kuangrong and Jin, Yaochu},
  title   = {A Survey of Evolutionary Algorithms for Multi-Objective Optimization Problems With Irregular Pareto Fronts},
  journal = {IEEE/CAA Journal of Automatica Sinica},
  year    = {2021},
  volume  = {8},
  number  = {2},
  pages   = {303--318},
  doi     = {10.1109/JAS.2021.1003817}
}

@inproceedings{Todorov_iLQR,
  title={Iterative Linear Quadratic Regulator Design for Nonlinear Biological Movement Systems},
  author={Weiwei Li and Emanuel Todorov},
  booktitle={International Conference on Informatics in Control, Automation and Robotics},
  year={2004}
}

@article{Todorov_motor_theory,
  author  = {Todorov, Emanuel and Jordan, Michael I.},
  title   = {Optimal Feedback Control as a Theory of Motor Coordination},
  journal = {Nature Neuroscience},
  year    = {2002},
  volume  = {5},
  number  = {11},
  pages   = {1226--1235}
}

\end{document}